\documentclass[conference]{IEEEtran}
\IEEEoverridecommandlockouts
\usepackage{cite}
\usepackage{amsmath,amssymb,amsfonts}
\usepackage{algorithmic}
\usepackage{graphicx}
\usepackage{textcomp}
\usepackage{comment}

\usepackage{booktabs}
\usepackage{multirow}
\usepackage{comment}
\usepackage{makecell}
\usepackage[svgnames]{xcolor}
\usepackage{subcaption}
\def\BibTeX{{\rm B\kern-.05em{\sc i\kern-.025em b}\kern-.08em
    T\kern-.1667em\lower.7ex\hbox{E}\kern-.125emX}}
\begin{document}

\title{Towards Fully Decoupled End-to-End Person Search\\
	\thanks{*Corresponding author.}
}

		\author{\IEEEauthorblockN{Pengcheng Zhang}
			\IEEEauthorblockA{\textit{School of Computer Science and Engineering, } \\
				\textit{State Key Laboratory of Software Development Environment, } \\
				\textit{Jiangxi Research Institute, Beihang University}\\
				Beijing, China \\
				pengchengz@buaa.edu.cn}
			\and
			\IEEEauthorblockN{Xiao Bai*}
			\IEEEauthorblockA{\textit{School of Computer Science and Engineering, } \\
				\textit{State Key Laboratory of Software Development Environment, } \\
				\textit{Jiangxi Research Institute, Beihang University}\\
				Beijing, China \\
				baixiao@buaa.edu.cn}
			\and
			\IEEEauthorblockN{Jin Zheng}
			\IEEEauthorblockA{\textit{School of Computer Science and Engineering, } \\
				\textit{State Key Laboratory of Software Development Environment, } \\
				\textit{Jiangxi Research Institute, Beihang University}\\
				Beijing, China \\
				jinzheng@buaa.edu.cn}
			\and
			\IEEEauthorblockN{Xin Ning}
			\IEEEauthorblockA{\textit{Institute of Semiconductors, Chinese Academy of Sciences,} \\
				\textit{Cognitive Computing Technology Joint Laboratory,}\\
				\textit{ Wave Group,}\\
				Beijing, China \\
				ningxin@semi.ac.cn}

		}

\maketitle

\begin{abstract}
End-to-end person search aims to jointly detect and re-identify a target person in raw scene images with a unified model. The detection task unifies all persons while the re-id task discriminates different identities, resulting in conflict optimal objectives. Existing works proposed to decouple end-to-end person search to alleviate such conflict. Yet these methods are still sub-optimal on one or two of the sub-tasks due to their partially decoupled models, which limits the overall person search performance. In this paper, we propose to fully decouple person search towards optimal person search. A task-incremental person search network is proposed to incrementally construct an end-to-end model for the detection and re-id sub-task, which decouples the model architecture for the two sub-tasks. The proposed task-incremental network allows task-incremental training for the two conflicting tasks. This enables independent learning for different objectives thus fully decoupled the model for persons earch. Comprehensive experimental evaluations demonstrate the effectiveness of the proposed fully decoupled models for end-to-end person search.
\end{abstract}

\begin{IEEEkeywords}
person search, decoupling, task-incremental learning 
\end{IEEEkeywords}

\section{Introduction}

Person search \cite{ps} jointly performs two sub-tasks, \textit{i.e.} person detection \cite{faster_rcnn,retina,fcos} and re-id \cite{bag_tricks,mgn,osnet,agw}, to locate a query person across a gallery of uncropped scene images. To guarantee the overall person search accuracy, it requires optimal detection performance to integrally contain all true positive persons, and optimal re-id performance to discriminate persons of different identities.

\begin{figure}[t]
	\centering
	\begin{subfigure}{0.48\textwidth}
		\includegraphics[width=\textwidth]{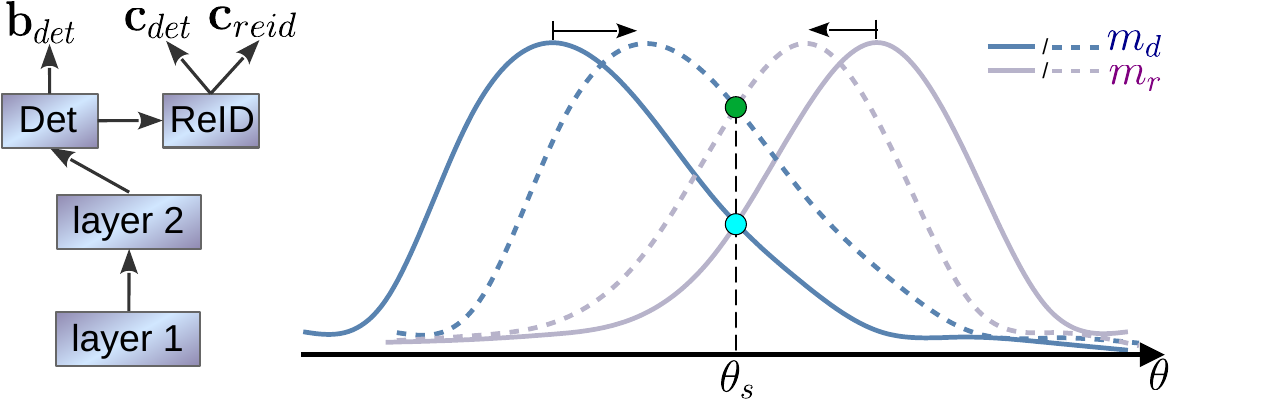}
		\caption{}
		\label{fig:cp}
	\end{subfigure}
	\begin{subfigure}{0.45\textwidth}
		\centering
		\includegraphics[width=\textwidth]{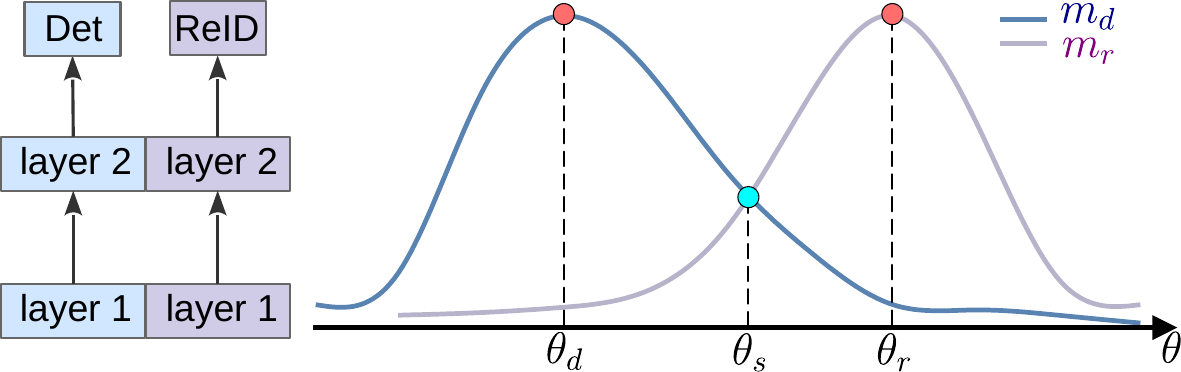}
		\caption{}
		\label{fig:ti}
	\end{subfigure}
	\caption{Comparison of fully decoupled person search (b) with previous decoupled models \cite{nae,hoim} (a). We employ \textcolor{Cyan}{cyan} points to indicate the performance of the vanilla end-to-end model \cite{oim}. \textbf{(a)} \textbf{Left:} Partially decoupled model in \cite{nae,hoim}. \textbf{Right:} The \textcolor{DarkGreen}{green} point indicates the performance of the partially decoupled model. By closing the two task-specific feature spaces, the upper bound of performance upon shared parameters $\mathbf{\theta}_s$ is boosted. \textbf{(b)} \textbf{Left:} The proposed fully decoupled person search network. \textbf{Right:} The \textcolor{Tomato}{pink} points illustrate the performance of the proposed model. It eliminates the coupled parameters and achieves the optimum for both sub-tasks towards optimal person search.}
	\label{fig:e2e1}
\end{figure}

Recent researches focus more on end-to-end methods \cite{oim,hoim,nae,seqnet,alignps,coat,pstr} that complete person search with a unified model. Although this paradigm shows better efficiency than the two-step ones \cite{prw,mgts,igpn,rdlr,tcts,btcl}, it suffers from the conflicting objectives of the two sub-tasks as the detection task unifies all persons while the re-id task discriminates different identities. For illustration purposes, we denote by $\mathbf{\theta}_s$ the model parameters shared the two tasks, and show the overall performance of a vanilla end-to-end model \cite{oim} in Figure \ref{fig:cp} where $m_d$ and $m_r$ denote the individual performance (larger is better) of the detection and re-id sub-task upon model parameters $\mathbf{\theta}$. Due to that $\mathbf{\theta}_s$ is the majority of model parameters, it obtains only a compromised average solution for person search. To alleviate this problem, previous works explored partially decoupled person search by closing the two task-specific feature spaces \cite{hoim,nae} and boosting the performance upper-bound upon $\mathbf{\theta}_s$ as in Figure \ref{fig:cp}. 

Another solution is to separate the respective prediction branches \cite{dmrnet,alignps} for different tasks, which constraints the impact of $\mathbf{\theta}_s$ and decouples the prediction parameters to improve sub-task performances.

Despite the progress made for decoupled person search, these methods are still sub-optimal on one or two of the sub-tasks due to the models are still partially coupled. This further limits the overall person search performances. To this end, we propose to fully decouple end-to-end models, as is shown in Figure \ref{fig:ti}, to achieve the optimum for both tasks towards the optimal solution for person search. The advantages of fully decoupled end-to-end person search are threefold: \textbf{(1)} The detection sub-network is comparable with standalone detectors, which maximumly reduces under-detected target persons in variant scene images; \textbf{(2)} Person retrieval features are learned without compromising to the detection sub-task, which guarantees the feature discrimination; \textbf{(3)} The architecture of the re-id sub-network is less dependent on the detection sub-network, which unleashes more space to design specialized re-id modules.

Inspired by the task-incremental learning (TIL) \cite{rusu2016progressive,yoon2018lifelong,side,wallingford2022task,ladder_side} mechanism, we propose a task-incremental network to enable the aforementioned fully decoupled end-to-end person search. With the assumption that the task identity is known during inference, TIL accumulates task-dependent parameters during training and selects proper ones at test time to mitigate catastrophic forgetting for incremental learning \cite{van2019three}. 
In contrast, for task-incremental person search, we construct an expandable model which is derived from a standared person detector and expanded for an incrmental task, \textit{i.e.} the re-id sub-task. The overall model architecture is thus decoupled for the two sub-tasks. 

The task-incremental person search network also degrades to a partially decoupled model when jointly trained by the two sub-tasks. We thus propose to conduct task-incremental training to decouple the training procedure. We first train a standard person detector and then expand the model for the re-id task. The whole model is then trained only by the re-id sub-task while the detection sub-network is frozen. The training for the re-id sub-task introduces spatial noises on GT boxes to simulate overlapping detection bounding boxes in previous works \cite{oim,hoim,nae,oim++,seqnet,coat} without performing computationally expensive person detection. Therefore, the model is fully decoupled for the two conflicting sub-tasks. This paradigm introduces an extra training phase, yet it keeps the end-to-end efficiency for inference. And a portion of the added re-id modules runs in parallel with the detection sub-network, which seldom increases the time cost.

In summary, this paper makes the following contributions:
\begin{itemize}
	\item We propose the first fully decoupled end-to-end person search model. By designing a novel task-incremental person search network, we decouple the model architecture for the two conflicting sub-tasks.
	
	\item The proposed task-incremental person search network allow task-incremental training for the two sub-tasks. This enables independent learning for the conflicting task objectives. Thus the proposed method achieves the optimum for the two conflicting sub-tasks towards optimal end-to-end person search.
	
	\item Comprehensive experiments demonstrate that the proposed model significantly outperforms previous decoupled models on PRW \cite{prw} and achieves the best on CUHK-SYSU \cite{oim}.
	
\end{itemize}

\section{Related Work}\label{rltw}

\begin{figure*}[htbp]
	\centering
	\begin{subfigure}{0.48\textwidth}
		\centering
		\includegraphics[scale=0.50]{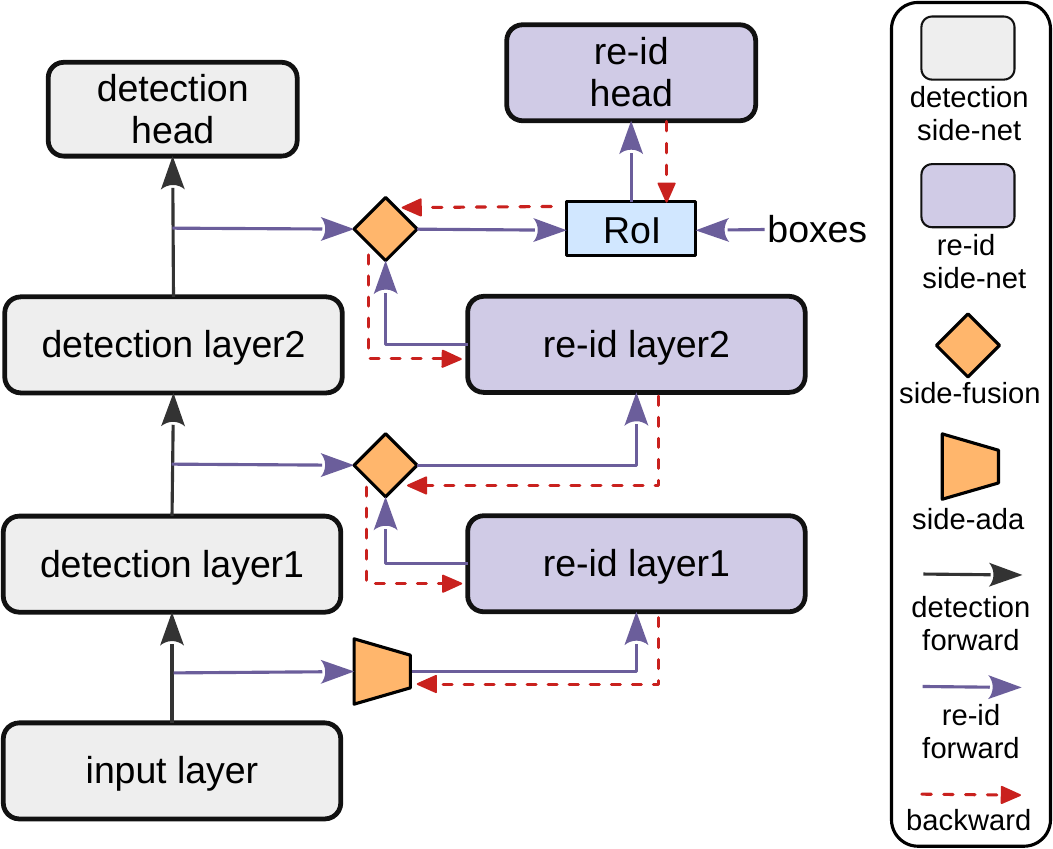}
		\caption{Overall architecture}
		\label{fig:tips}
	\end{subfigure}
	\begin{subfigure}{0.48\textwidth}
		\centering
		\includegraphics[scale=0.50]{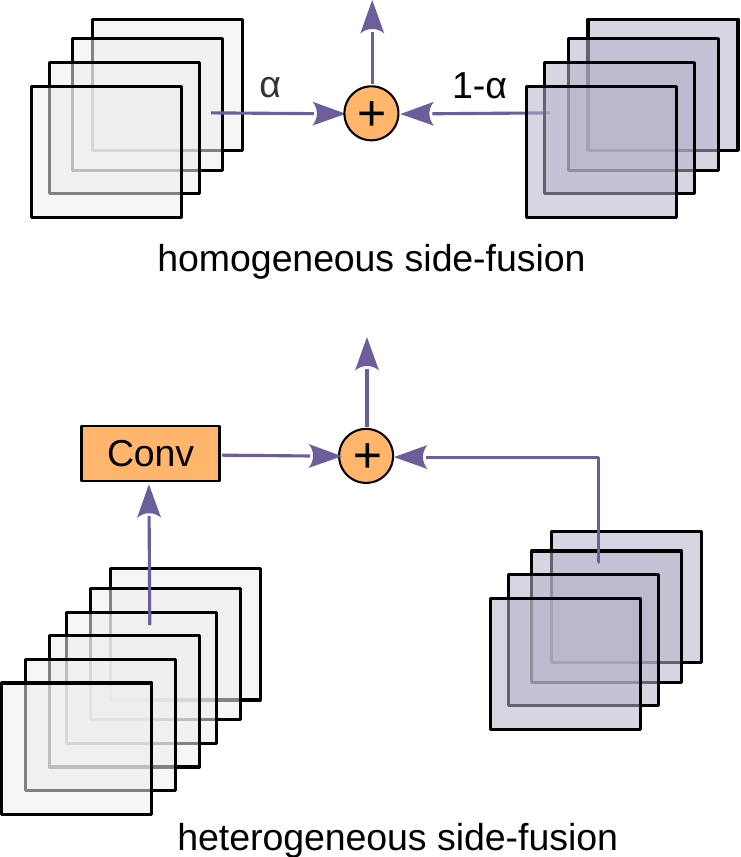}
		\caption{The \emph{side-fusion} module}
		\label{fig:fuse}
	\end{subfigure}
	
	\caption{(a) The proposed fully decoupled person search network which consists of a \emph{detection side-net}, a \emph{re-id side-net}, and the modules that bridge them. The model is trained incrementally by person detection and re-id tasks, which fully decouples the parameters for the two conflicting sub-tasks. (b) The architectures of homogeneous and heterogeneous \emph{side-fusion}s for their corresponding \emph{re-id side-net}. These modules transfer knowledge from the trained \emph{detection side-net} to the \emph{re-id side-net}.}
	\label{fig:model}
	
\end{figure*}
\textbf{Person Search.} Person search methods can be categorized into two types: two-step methods and end-to-end methods. Two-step methods typically employ a standalone detector to detect and crop person images, and a re-id model to retrieve a target across the cropped images. \cite{prw} first explores combining popular person detector and re-id models for person search. To obtain more representative features for each identity, \cite{mgts} proposes to enhance foreground image regions and \cite{clsa} performs multi-scale matching for person search. \cite{igpn,tcts,btcl} instead design target-guided person detectors to suppress retrieval distractors. 
To improve the efficiency of the two-step paradigms, end-to-end methods propose to perform person search by a unified model. \cite{oim} constructs the first end-to-end model and proposes the Online Instance Matching (OIM) \cite{oim} loss for model training.
Following \cite{oim}, \cite{qeeps,ctxg,seqnet,pga} significantly improve the performances of end-to-end models. \cite{pstr,coat} draw inspiration from the Transformers \cite{vit,ddetr} and obtain more discriminative person features with well-designed models.

One key challenge for end-to-end person search is the contradictory objectives of person detection and re-id. To deal with the problem, \cite{hoim} proposes hierarchical classification score calculation and \cite{nae} disentangles the output embedding for the two tasks. These close the gap between two marginal task-specific feature spaces. Another solution is to construct separate prediction branches as in \cite{dmrnet}, which decouples the predictions for different tasks. \cite{alignps} proposes a "re-id first" design and can be viewed as employing a non-parametric identity re-id prediction branch for this. Though achieved significant progress, these models are still sub-optimal on one or two of the sub-tasks due to that a subset of model parameters are left coupled for the two sub-tasks. This limits the overall person search performances of the models. To this end, this paper proposes to fully decouple end-to-end models towards optimal person search.

\textbf{Task-incremental Learning.} TIL \cite{van2019three} is one of the major scenarios for incremental learning. A key assumption for TIL is that the task identity is known during inference. The catastrophic forgetting problem is thus can be solved by incrementally learning task-specific sub-networks while preserving all learned ones. At test time, the task identity is then employed to select the proper sub-network. Specifically, \cite{rusu2016progressive,li2019learn,wang2017growing,yoon2018lifelong} directly expands the network by adding new layers or branches for new tasks. Yet this can be limited in practice due to unbounded parameter growth. Other works thus propose to freeze partial network with masks for old tasks \cite{golkar2019continual,hung2019compacting,mallya2018packnet,serra2018overcoming} and adapt to the new task with left trainable parameters, which may suffer from running out of model capacities. To mitigate the limitation that the task identity is not always available, \cite{abati2020conditional,wallingford2022task} additionally trains a task classifier to infer task identity or designs adaptive parameter selection mechanisms at test time. With similar techniques, \cite{side,ladder_side} also suits TIL by incrementally expandable side-networks.

In this work, we design a task-incremental network to enable the aforementioned fully decoupled end-to-end person search. This achieves the optimum for both the two sub-tasks to facilitate optimal person search performance.
\section{Method }
\subsection{Task-incremental Person Search Network }\label{sec:tips}

\textbf{Detection side-net.} The \emph{detection side-net} $f_d$ can be various modern detectors, \textit{e.g.} Faster R-CNN \cite{faster_rcnn}, RetinaNet \cite{retina} and FCOS \cite{fcos}, given the simple and expandable overall architecture. For illustration purposes, we employ the Faster R-CNN \cite{faster_rcnn} detector as $f_d$ in this section. The \emph{input layer} in Figure \ref{fig:tips} is composed of the `conv1' and `conv2' blocks of the ResNet \cite{resnet} backbone. The \emph{detection layer1} and \emph{detection layer2} are the `conv3' and `conv4' blocks, respectively. The \emph{detection head} consists of the RPN \cite{faster_rcnn} and `conv5' block to predict probable person locations and corresponding classification scores. Following the common practice in object detection \cite{faster_rcnn,retina,fcos,ddetr}, the \emph{input layer} is initialized with ImageNet \cite{imagenet} pre-trained parameters and frozen during training, which makes it independent neither on the detection sub-task nor on the re-id sub-task.

\begin{figure}[h]
	\centering
	\includegraphics[width=0.45\textwidth]{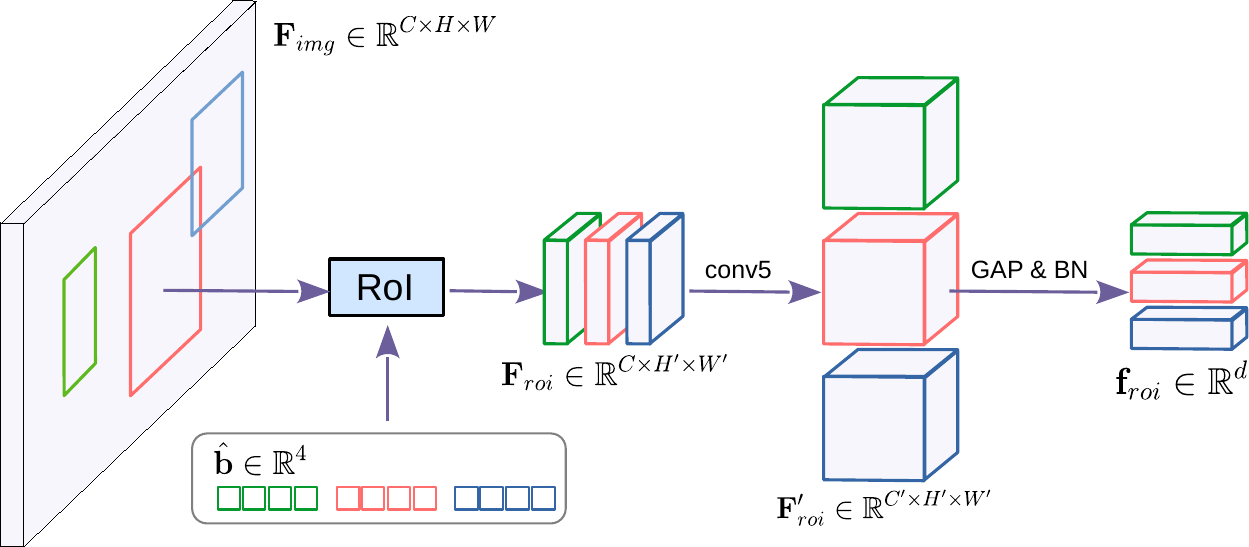}
	\caption{Illustration of the \emph{re-id head}. Person feature maps are drawn from the output of `conv4' and refined by the `conv5' block. By consecutive global average pooling and batch normalization, this module produces 1-D person feature vectors.}
	\label{fig:reid_head}
	
\end{figure}

\textbf{Re-id side-net.} As is in Figure \ref{fig:tips}, the \emph{re-id side-net} $f_r$ shares the \emph{input layer} with $f_d$. Similar to $f_d$, another `conv3' block and `conv4' block are employed as the consecutive \emph{re-id layer}s to extract image feature maps $\mathbf{F}_{img} \in \mathbb{R}^{C\times H\times W}$. Given probable person bounding boxes $\hat{\mathbf{b}}\in \mathbb{R}^4$, as is in Figure \ref{fig:reid_head}, RoIAlign \cite{mask_rcnn} is performed to obtain corresponding person feature maps $\mathbf{F}_{roi} \in \mathbb{R}^{C\times H' \times W'}$. We further employ the `conv5' block (stride is set to $1$) of the backbone on top of $\mathbf{F}_{roi}$ to produce $\mathbf{F}'_{roi} \in \mathbb{R}^{C'\times H' \times W'}$. Subsequent global average pooling (GAP) and batch normalization (BN) \cite{bn} layers then aggregate $\mathbf{F}'_{roi}$ to 1-D person features $\mathbf{f}_{roi} \in \mathbb{R}^{d}$ for person retrieval.

\textbf{Side-ada.} The network architecture of $f_r$ can be homogeneous (\textit{e.g.}, ResNet50 for both ) or heterogeneous (\textit{e.g.}, ResNet50 for $f_d$ and ResNet34 for $f_r$) to $f_d$ . The \emph{side-ada} module is thus inserted to transform the output of the \emph{input layer} to match the input size required by $f_r$. For homogeneous \emph{re-id side-net}, the \emph{side-ada} module is implemented by an identity function $I\left(\cdot\right)$ where

\begin{equation}
	I\left(\mathbf{x}\right)=\mathbf{x}, \mathbf{x} \in \mathbb{R}^{c'\times h' \times w'}.
\end{equation}

And for heterogeneous \emph{re-id side-net}, we instead employ a single convolution layer \\

\begin{equation}
	Conv\left(\mathbf{x}\right)=\mathbf{x}', 
	\vspace{-1mm}
\end{equation}
where $\mathbf{x} \in \mathbb{R}^{c\times h \times w}$ and $\mathbf{x}' \in \mathbb{R}^{c'\times h' \times w'}$, to transform the input to be spatially compatible with $f_r$.

\textbf{Side-fusion.} Inspired by \cite{side,ladder_side}, we further add the \emph{side-fusion} module to fuse $\mathbf{x}_{d}^{i}$ and $\mathbf{x}_{r}^{i}$, the outputs of two $i$th parallel side-network layers, as input $\mathbf{x}_{s}^{i}$ to the subsequent \emph{re-id side-net} block. This transfers valuable knowledge from the pre-trained $f_d$ to $f_r$. Analogously, we design homogeneous \emph{side-fusion} and heterogeneous \emph{side-fusion} modules for homogeneous $f_r$ and heterogeneous $f_r$, respectively. As Figure \ref{fig:fuse} shows, the homogeneous \emph{side-fusion} performs alpha blending

\begin{equation}
	\mathbf{x}_{s}^{i}=\alpha_i\mathbf{x}_{d}^{i}+\left(1-\alpha_i\right)\mathbf{x}_{r}^{i},
\end{equation}
where $\mathbf{x}_{s}^{i}$, $\mathbf{x}_{d}^{i}$ and $\mathbf{x}_{r}^{i}$ are all in $\mathbb{R}^{c'_i\times h'_i\times w'_i}$, to fuse the outputs. $\alpha_i \in \left[0,1\right]$ is the only learnable parameter of this module. Similar to the \emph{side-ada} module, the heterogeneous \emph{side-fusion} module employs a single convolution layer to transform $\mathbf{x}_{d}^{i}$ to be spatially compatible with $\mathbf{x}_{r}^{i}$. The fused output is thus given by

\begin{equation}
	\mathbf{x}_{s}^{i}=Conv\left(\mathbf{x}_{d}^{i}\right)+\mathbf{x}_{r}^{i}
\end{equation}
where $\mathbf{x}_{s}^{i}$, $\mathbf{x}_{r}^{i} \in \mathbb{R}^{c'_i\times h'_i\times w'_i}$ and $\mathbf{x}_{d}^{i} \in \mathbb{R}^{c_i\times h_i\times w_i}$. 

\subsection{Task-incremental Model Training}
The proposed task-incremental person search network in Section \ref{sec:tips} is still partially decoupled when conducting simple joint training for person search. To this end, we propose task-incremental training for the task-incremental person search network. Specifically, we first train a standard person detector $f_d$ on the training set with detection losses according to the specific settings of the detector. Afterwards, we freeze the detection side network $f_d$ and expand the model with $f_r$ and the bridge modules. The whole model is then trained only by the re-id task with the OIM loss \cite{oim} $\mathcal{L}_{oim}$ and the triplet loss $\mathcal{L}_{tri}$ similar to \cite{alignps}, which makes $f_r$ learn discriminative person representations without competing with $f_d$. The overall model is thus fully decoupled in a task-incremental learning manner for person search. 

\begin{figure}[h]
	\centering
	\includegraphics[width=0.39\textwidth]{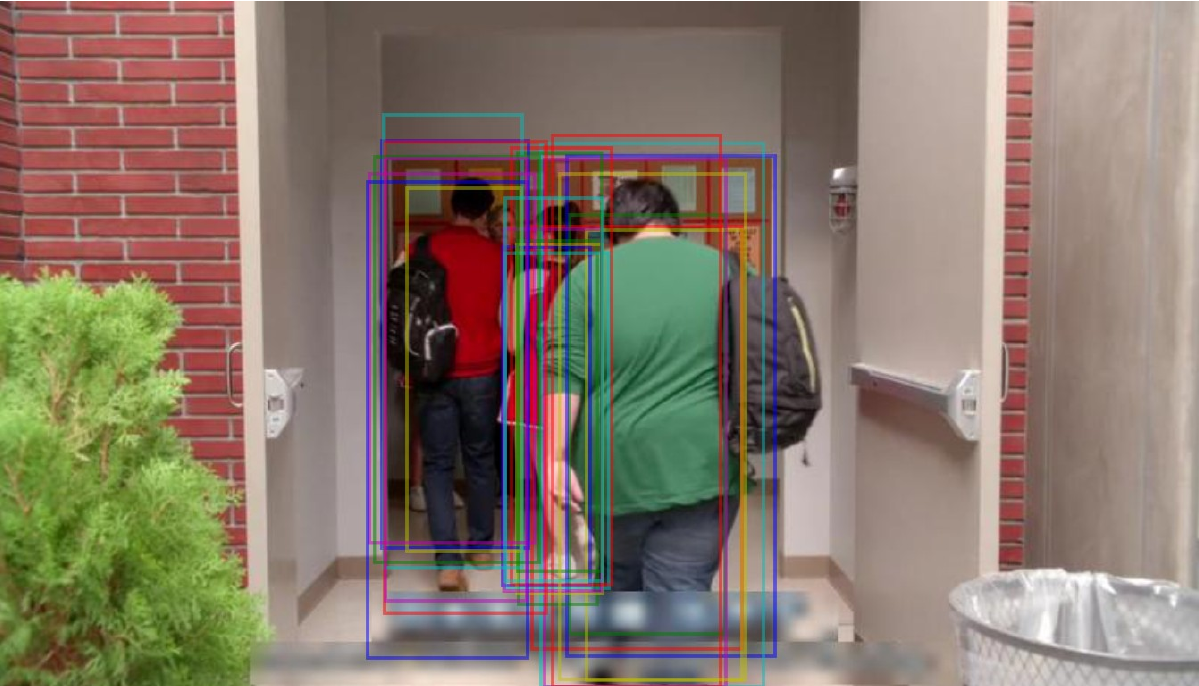}
	\caption{Illustration of augmented person bounding boxes from the GT boxes. Center shifting and box scaling are employed to add spatial noises.}
	\label{fig:box_aug}
\end{figure}

Note that previous works \cite{oim,nae,alignps,seqnet,coat} detect multiple overlapping bounding boxes of the same person and extract feature embeddings upon the boxes to train the re-id modules, which implicitly peforms spatial augmentation for representation learning. To decouple the training of the re-id side-network from the detection side-network without sacrificing this advantage, we peopose to add spatial noises, namely Spatial-noise Augmentation (SnA), to the GT boxes to obtain augmented training samples for $f_r$. Similar to \cite{dn_detr}, we combine two types of spatial noises: center shifting and box scaling. Denoted by $\left(c_x,c_y\right)$ and $\left(h_{\mathbf{b}},w_{\mathbf{b}}\right)$ the center point and spatial size of GT box $\mathbf{b}$, center shifting adds random noise $\left(\Delta c_x,\Delta c_y\right)$ to the center, where $|\Delta c_x|<\frac{\lambda_1 w_{\mathbf{b}}}{2}$ and $|\Delta c_y|<\frac{\lambda_1 h_{\mathbf{b}}}{2}$. And box scaling randomly samples height and width from $\left[\left(1-\lambda_2\right)h_{\mathbf{b}},\left(1+\lambda_2\right)h_{\mathbf{b}}\right]$ and $\left[\left(1-\lambda_2\right)w_{\mathbf{b}},\left(1+\lambda_2\right)w_{\mathbf{b}}\right]$. $\lambda_1\in \left(0,1\right)$ and $\lambda_2\in \left(0,1\right)$ are hyper-parameters that control the noise scales of the augmentations. For training, as in Figure \ref{fig:box_aug}, we randomly generate $n$ augmented versions of each GT box to enhance the robustness of $f_r$.

\section{Experiments}\label{exprs}
\subsection{Datasets}
\textbf{CUHK-SYSU} \cite{oim} presents 18,184 images from both movies and street snapshots. A total number of 96,143 pedestrian bounding boxes and 8,432 labeled identities are manually annotated for person search. The training subset provides 11,206 frames with 5532 identities from both sources. The testing subset selects 2900 query identities with gallery sizes varying from 50 to 4,000. The training and testing sets have no overlaps on images or identities.

\textbf{PRW} \cite{prw} deployed 6 cameras at a campus to record multi-view videos of pedestrians. 11,816 frames are selected and densely annotated as 43,110 pedestrian bounding boxes with 932 identities. The training subset contains 5,134 frames with 432 identities, while the rest 6,112 frames make up the test subset. The evaluation protocol of PRW takes the full test set as the gallery for evaluation by default, which tends to be more challenging than CUHK-SYSU.

\textbf{Evaluation Metrics}. Similar to the evaluation metrics in person re-identification \cite{market1501,mtmc}, the mAP and top-1 accuracy are utilized in person search. During the evaluation, the gallery of probable persons is dynamically built upon detection results. A retrieved person is considered positive if it shares the same identity with the query and the detected bounding box has the Intersection over Union (IOU) larger than 0.5 with the GT, which makes the mAP and top-1 accuracy also be affected by person detection results. For the evaluation of person detection, the Average Precision (AP) and Recall \cite{Everingham10} are the mostly employed metrics in recent person search works. 
\subsection{Implementation Details}

For training, we first train $f_d$ by person detection task for $18$ epochs with batch size of $8$. Then we freeze the parameters of $f_d$ and train the whole model by the re-id sub-task for $18$ epochs with batch size of $5$.
The output size of RoIAlign is set to $16\times 8$ for both $f_d$ and $f_r$. The input image is randomly scaled for training and resized to $1500\times 900$ for testing. We use SGD optimizer with initial learning rate of $0.003$. The learning rate is linearly warmed up during the first epoch. For $f_d$, we decrease the learning rate by $10$ at the $12^{th}$ epoch. And for $f_r$, the learning rate is decreased by $10$ at the $10^{th}$ epoch. For CUHK-SYSU/PRW, the circular queue size of OIM is set to 5000/500 and the softmax temperature is set to $1/30$.

In the following experiments, we employ Faster R-CNN with ResNet50 backbone as $f_d$ and ResNet50 blocks initialized with ImageNet pre-trained parameters as $f_r$ unless otherwise specified. We use $^{\dagger}$ to mark models evaluated with Class Weighted Similarity (CWS) \cite{prw}. The subscript $_{34}$ indicates that ResNet34 blocks are employed for $f_d$ and $f_r$.  Evaluations on CUHK-SYSU are with gallery size of 100 by default. The models are all trained and tested on a single RTX 3090 GPU. Extensive experimental results and implementation details are further presented in the Supplementary Material.

\subsection{Analytical Studies}

\begin{table}[h]
	\centering
	\begin{tabular}{l|cc|cc}
		\toprule
		\multirow{2}*{\textbf{Detectors}} & \multicolumn{2}{c|}{\textbf{PRW}} & \multicolumn{2}{c}{\textbf{CUHK-SYSU}} \\
		\cline{2-5}
		& \textbf{mAP} & \textbf{top-1} & \textbf{mAP} & \textbf{top-1} \\
		\midrule
		RetinaNet \cite{retina} & 47.6 & 86.3 & 92.8 &93.8\\
		FCOS \cite{fcos} & \underline{47.8}  & \underline{86.7} & \underline{93.0} &\textbf{94.3}\\
		Fatser R-CNN \cite{faster_rcnn} & 47.9 & 86.5 & 92.1 & 93.4\\
		\midrule
		RetinaNet$^{\triangledown}$ \cite{retina} & 22.1 & 55.8 & 73.6 & 75.2\\
		FCOS$^{\triangledown}$ \cite{fcos} & 22.4  & 56.4 & 76.9 & 77.8 \\
		Faster R-CNN$^{\triangledown}$ \cite{faster_rcnn} & 21.0 & 47.2 & 74.2 & 77.6 \\
		\midrule
		RetinaNet$^{\dagger}$ \cite{retina} & 46.8 & 85.9 & 93.1 & 93.8\\
		FCOS$^{\dagger}$ \cite{fcos} &  46.1 & 85.5 & 92.6 & 92.9\\
		Faster R-CNN$^{\dagger}$ \cite{faster_rcnn} & \textbf{49.0} & \textbf{87.1} & \textbf{93.4} & \underline{94.2}  \\
		\bottomrule
	\end{tabular}
	\caption{The person search evaluation results of our proposed model with different detectors as $f_d$. We use $^{\triangledown}$ to denote the coupled models. }
	\label{tab:dets}
	
\end{table}
\textbf{Different choices of $f_d$.} The proposed fully decoupled person search network specifies no concrete architecture of $f_d$. To verify the effect of different detector architectures, we employ RetinaNet \cite{retina}, FCOS \cite{fcos} and Faster R-CNN \cite{faster_rcnn}, as $f_d$ for our proposed model and present the evaluation results in Table \ref{tab:dets}. On PRW, the proposed model achieves similar performance with different detectors. And on CUHK-SYSU, employing RetinaNet or FCOS as $f_d$ slightly outperforms that with Faster R-CNN. These results demonstrate that the proposed fully decoupled network is well-compatible with various detector architectures. We additionally test their respective coupled versions, \textit{i.e.}, models jointly trained by the two sub-tasks and directly share the features between the detection prediction layers and the re-id losses. The proposed fully decoupled models also outperform the coupled versions by a large margin. We further conduct the evaluation with CWS \cite{prw} on the models as in \cite{nae,seqnet}. This significantly boosts the person search performance when employing Faster R-CNN as $f_d$ yet harms that with other detectors. We present a more comprehensive analysis upon this in the Supplementary Material.

\begin{table}[h]
	\centering
	\begin{tabular}{l|cc|cc}
		\toprule
		\textbf{Combination of  } & \multicolumn{2}{c|}{\textbf{PRW}} & \multicolumn{2}{c}{\textbf{CUHK-SYSU}} \\
		\cline{2-5}
		\textbf{$f_d$ and $f_r$}& \textbf{mAP} & \textbf{top-1} & \textbf{mAP} & \textbf{top-1} \\
		\midrule
		R-50 \& R-50  & \textbf{47.9} & \textbf{86.5} & \textbf{92.1} & \underline{93.4}\\
		R-34 \& R-34 & \underline{47.1} &\underline{ 85.5} & \underline{92.0} & \textbf{93.3} \\
		\midrule
		R-50 \& R-34 & 41.6 & 83.7 & 91.0 & 92.1 \\
		R-50 \& OSNet \cite{osnet} & 40.9 & 84.7 & 88.6 & 90.2\\
		\bottomrule
	\end{tabular}
	\caption{Comparison between different combinations of $f_d$ and $f_r$. The first two rows are homogeneous $f_r$s while others are heterogeneous $f_r$s. We use R-50 and R-34 to denote ResNet50 and ResNet34, respectively.}
	\label{tab:hh}
	
\end{table}
\textbf{Comparison between homogeneous $f_r$ and heterogeneous $f_r$.} As is described in Section \ref{sec:tips}, $f_r$ can be with homogeneous or heterogeneous architecture to $f_d$. 
To investigate the effects of different combinations of $f_d$ and $f_r$, we conduct experiments as in Table \ref{tab:hh}. We initialize the $f_r$s with their respective ImageNet pre-trained parameters for fair comparison. It can be observed that the performances of the heterogeneous combinations are marginally inferior to the homogeneous counterparts. The main reason for this can be the limited capacity of $f_r$ or the misalignment between heterogeneous blocks. We thus additionally test a homogeneous combination of ResNet34 for both $f_d$ and $f_r$. Although the capacity of this combination is also limited, it only performs moderately inferior to the ResNet50 combination, which demonstrates that the misalignment between heterogeneous blocks mainly impedes the performance. We thus employ homogeneous side-networks in the proposed model by default.

	\begin{table}[h]
		\centering
		\begin{tabular}{l|cc|cc}
			\toprule
			\multirow{2}*{\textbf{Models}} & \multicolumn{2}{c|}{\textbf{PRW}} & \multicolumn{2}{c}{\textbf{CUHK-SYSU}} \\
			\cline{2-5}
			& \textbf{mAP} & \textbf{top-1} & \textbf{mAP} & \textbf{top-1} \\
			\midrule
			ours w/ \emph{side-fusion} & \textbf{47.9} & \textbf{86.5} & \textbf{92.1} & \underline{93.4}\\
			ours w/o \emph{side-fusion} & 46.1 & 85.5 & 91.1 & 92.4 \\
			ours$_{34}$ w/ \emph{side-fusion} & \underline{47.1} & \underline{85.5} & \underline{92.0} & \textbf{93.3}\\
			ours$_{34}$ w/o \emph{side-fusion} & 44.4 & 84.7 & 91.0 & 92.1 \\
			\bottomrule
		\end{tabular}
		\caption{Person search performances of the proposed models with and without \emph{side-fusion}. Note that the tested \emph{side-fusion} modules are homogeneous \emph{side-fusion} modules.}
		\label{tab:fusion}
		
	\end{table}
	\textbf{The effect of \emph{side-fusion}.} We propose the \emph{side-fusion} modules to transfer useful knowledge, \textit{e.g.} suppressing the background and enhancing the foreground, from the trained $f_d$ to $f_r$. The proposed model is also capable of performing person search without the \emph{side-fusion} modules. We thus evaluate the person search performances of models with and without \emph{side-fusion}, as in Table \ref{tab:fusion}, to verify the impact of the modules. It can be observed that the \emph{side-fusion} modules consistently boost the person search accuracy, suggesting that the knowledge learned from the detection sub-task can be implicitly employed to strengthen the re-id side-network.

\textbf{Comparison between joint training and task-incremental training}. To enable the fully decoupled person search, we propose a task-incremental person search network and train the model in a task-incremental manner. It is worth noting that the proposed task-incremental person search network becomes a partially decoupled model when jointly trained for the two sub-tasks. Specifically, the shared \textit{input layer} is fixed with ImageNet pre-trained parameters and thus independent of the two sub-tasks. The \textit{side fusion} modules are extremely lightweight, which constraints the coupled optimization for the contradictory objectives. And the prediction modules are separated for their respective tasks similar to \cite{dmrnet}. To validate the effect of task-incremental training, we conduct performance comparisons between joint training and task-incremental training. As the results in Table \ref{tab:training_ps} and Table \ref{tab:training_det} show, the model also obtains comparable performances to previous decoupled methods when jointly trained by the two sub-tasks. And task-incremental training further boosts the person search performance as well as the person detection performance. We also test a hybrid training scheme, \textit{i.e.} jointly training $f_d$ with a very small $lr=1.0\times 10^{-5}$ when training $f_r$. The results in Table \ref{tab:training_ps} and Table \ref{tab:training_det} suggest that this slightly improves the person search performance yet significantly impedes the person detection capability. And the memory cost is also increased when jointly training the two side-networks.

\begin{table}[h]
	\centering
	\begin{tabular}{l|cc|cc}
		\toprule
		\multirow{2}*{\textbf{Training}} & \multicolumn{2}{c|}{\textbf{PRW}} & \multicolumn{2}{c}{\textbf{CUHK-SYSU}} \\
		\cline{2-5}
		& \textbf{mAP} & \textbf{top-1} & \textbf{mAP} & \textbf{top-1} \\
		\midrule
		joint & 47.3 & 86.1 & 91.3 & 92.7 \\
		task-incremental & 47.9 & 86.5 & 92.1 & 93.4 \\
		hybrid & 48.0 & 86.2 & 92.2 & 93.5 \\
		\bottomrule
	\end{tabular}
	\caption{Person search performance comparison between joint training and task-incremental training.}
	\label{tab:training_ps}
\end{table}
\begin{table}[h]
	\centering
	\begin{tabular}{l|cc|cc}
		\toprule
		\multirow{2}*{\textbf{Training}} & \multicolumn{2}{c|}{\textbf{PRW}} & \multicolumn{2}{c}{\textbf{CUHK-SYSU}} \\
		\cline{2-5}
		& \textbf{AP} & \textbf{Recall} & \textbf{AP} & \textbf{Recall} \\
		\midrule
		joint & 92.8 & 96.8 & 87.0 & 93.3\\
		task-incremental & 93.4 & 97.6 & 87.8 & 94.0 \\
		hybrid & 90.4 & 95.1 & 82.7 & 86.1 \\
		\bottomrule
	\end{tabular}
	\caption{Person detection performance comparison between joint training and task-incremental training.}
	\label{tab:training_det}
\end{table}

\subsection{Comparison with State-of-the-art}
\begin{table}[h]
	\centering
	\begin{tabular}{l|cc|cc}
		\toprule
		\multirow{2}*{\textbf{Method}} & \multicolumn{2}{c|}{\textbf{PRW}} & \multicolumn{2}{c}{\textbf{CUHK-SYSU}} \\
		\cline{2-5}
		& \textbf{mAP} & \textbf{top-1} & \textbf{mAP} & \textbf{top-1} \\
		\midrule
		HOIM \cite{hoim} & 39.8 & 80.4 & 89.7 & 90.8 \\
		NAE$^{\dagger}$ \cite{nae} & 43.3  & 80.9 & 91.5 & 92.4\\
		NAE+$^{\dagger}$ \cite{nae} & 44.0 & 81.1 & 92.1 & 92.9 \\
		ours$_{34}^{\dagger}$  & \underline{47.3} & \underline{85.9} & \underline{92.6} & \underline{93.4} \\
		ours$^{\dagger}$ & \textbf{49.0} & \textbf{87.1} & \textbf{93.4} & \textbf{94.2} \\
		\midrule
		DMRNet \cite{dmrnet} & 46.1 & 83.2 & 91.6 & 93.0 \\
		ours$_{34}$ (RetinaNet) & \underline{46.5} & \underline{84.9} & \underline{92.0} & \underline{93.1} \\
		ours (RetinaNet) & \textbf{47.6} &\textbf{86.3} & \textbf{92.8} & \textbf{93.8} \\
		\midrule
		AlignPS \cite{alignps} & 45.9 & 81.9 & 93.1 & 93.4 \\
		AlignPS+ \cite{alignps} & 46.1 & 82.1 & \textbf{94.0} & \textbf{94.5} \\
		ours$_{34}$ (FCOS) & \underline{46.3} & \underline{85.4} & 92.6 & 93.4 \\
		ours(FCOS) & \textbf{47.8} & \textbf{86.7} & \underline{93.0} & \textbf{94.3} \\
		\bottomrule
	\end{tabular}
	
	\caption{Person search performance comparison with existing decoupled person search methods. We employ the results with RetinaNet of \cite{dmrnet} for fair comparison.}
	\label{tab:decouple_ps}
	
\end{table}

To demonstrate the advantages of fully decoupled person search networks, we first compare our proposed model with existing decoupled person search models in terms of person search (Table \ref{tab:decouple_ps}) and person detection (Table \ref{tab:decouple_det}) performances. We then present the comparison results with recent well-established state-of-the-art methods (Table \ref{tab:sota}). And qualitative visualization is shown at last.

\textbf{Results on CUHK-SYSU.} As is shown in Table \ref{tab:decouple_ps}, with similar configurations of detectors, our proposed models significantly outperform that in \cite{hoim,nae,dmrnet,alignps} on CUHK-SYSU. By applying decoupled model initialization, the proposed model achieves the best top-1 accuracy and mAP score. Note that AlignPS+ \cite{alignps} utilizes deformable convolution backbone \cite{dcn} and multi-scale features which are not included in our models. We further evaluate the detection performances of previous decoupled models and our proposed ones. By fully decoupling for the detection and re-id sub-tasks, the \emph{detection side-net} $f_d$ inherits the capability of independently trained detectors. The detection sub-network of our proposed model thus guarantees the optimum for person detection and significantly outperforms previous methods. When compared with recent well-established models, as in Table \ref{tab:sota}, the proposed model achieves competitive mAP and top-1 scores without bells and whistles. Compared with the best model COAT \cite{coat} that employs self-attention blocks \cite{vit} and cascaded refinement \cite{cascade} of multi-scale features, the proposed model achieves comparable performance with simple model architecture and single-scale features. We also test the combination of our proposed full decoupled perosns earch framework with the modules in \cite{coat}. This achieves superior performances on CUHK-SYSU, demonstrating the effectiveness of pur proposed method.

\begin{table}[h]
	\centering
	\begin{tabular}{l|cc|cc}
		\toprule
		\multirow{2}*{\textbf{Detector}} & \multicolumn{2}{c|}{\textbf{PRW}} & \multicolumn{2}{c}{\textbf{CUHK-SYSU}} \\
		\cline{2-5}
		& \textbf{AP} & \textbf{Recall} & \textbf{AP} & \textbf{Recall} \\
		\midrule
		Faster R-CNN by \cite{hoim} & 87.8 & 95.7 & 85.7 & 91.8 \\
		Faster R-CNN by \cite{nae} & 88.8  & 93.3 & 86.8 & 92.6\\
		ours Faster R-CNN & \textbf{93.4} & \textbf{97.6} & \textbf{87.8} & \textbf{94.0} \\
		\midrule
		RetinaNet by \cite{dmrnet} & - & - & 91.3 & - \\
		ours RetinaNet & 93.0 & 95.6 & \textbf{91.7} &  97.5  \\
		\midrule
		FCOS by \cite{alignps} & 88.4 & 90.5 & 86.9 & 89.1 \\
		FCOS$^{+}$ by \cite{alignps} & 89.1 & 91.1 & 86.0 & 88.8 \\
		ours FCOS & \textbf{93.4} & \textbf{95.8} & \textbf{92.2} &  \textbf{95.5}  \\
		\bottomrule
	\end{tabular}
	
	\caption{Person detection performance comparison with existing decoupled person search methods. $^{+}$ indicates the model with DCN \cite{dcn} backbone.}
	\label{tab:decouple_det}
	
\end{table}
\begin{table}[h]
	\centering
	\begin{tabular}{l|l|cc|cc}
		\toprule
		\multicolumn{2}{c|}{} & \multicolumn{2}{c|}{\textbf{PRW}} & \multicolumn{2}{c}{\textbf{CUHK-SYSU}} \\
		\cline{3-6}
		\multicolumn{2}{c|}{\multirow{-2}{*}{\textbf{Methods}}}& \textbf{mAP} & \textbf{top-1} & \textbf{mAP} & \textbf{top-1} \\
		\midrule
		\multirow{6}{*}{ \rotatebox{90}{two-step}} &IDE$^\dagger$ \cite{prw} & 20.5 & 48.3 & - & - \\
		&MGTS \cite{mgts} & 32.6 & 72.1 & 83.0 & 83.7 \\
		&CLSA \cite{clsa} & 38.7 & 65.0 & 87.2 & 88.5 \\
		&RDLR \cite{rdlr} & 42.9 & 70.2 & 93.0 & 94.2 \\
		&IGPN \cite{igpn} &\underline{47.2} & 87.0 & 90.3 & 91.4 \\
		&TCTS \cite{tcts} & 46.8 &\underline{87.5} &\underline{93.9} &\underline{95.1} \\
		\midrule
		\midrule
		\multirow{12}{*}{ \rotatebox{90}{end-to-end}} &OIM \cite{oim}  & 21.3 & 49.4 & 75.5 & 78.7 \\
		&CTXG \cite{ctxg}& 33.4 & 73.6 & 86.5 & 84.1 \\
		&BINet \cite{binet} & 45.3 & 81.7 & 90.0 & 90.7 \\
		&PGA \cite{pga}& 44.2 & 85.2 & 92.3 & 94.7 \\
		&SeqNet$^{\dagger}$ \cite{seqnet}& 45.8 & 81.7 & 93.4 & 94.1 \\
		&OIMNet++ \cite{oim++} & 47.7 & 84.8 & 93.1 & 94.1 \\
		&PSTR \cite{pstr}(R-50)& 49.5 & \textbf{87.8} & 93.5 & \underline{95.0} \\
		&COAT$^{\dagger}$ \cite{coat}& \underline{53.3} &87.4 & \underline{94.2} & 94.7 \\
		&ours$^{\dagger}$ & 49.0 & 87.1 & 93.4 & 94.2\\
		&ours + COAT$^{\dagger}$ \cite{coat} & \textbf{54.0} & \underline{87.5} & \textbf{94.3} & \textbf{95.2} \\
		\bottomrule
	\end{tabular}
	
	\caption{Comparison with other state-of-the-art methods. We also evaluate the models on the multi-view gallery of PRW as in \cite{coat} and present the results in the last 4 rows.}
	\label{tab:sota}
	
\end{table}

\textbf{Results on PRW.} On the PRW dataset, as Table \ref{tab:decouple_ps} shows, our proposed models surpass the best of previous decoupled methods by a large margin even with lightweight ResNet34 blocks. We also test the detection performances of models in \cite{hoim,nae,alignps} by their released checkpoints and present the results in Table \ref{tab:decouple_det}. It can be observed that our proposed models consistently achieve better performances than these methods. When compared with recent state-of-the-art end-to-end methods, the proposed model achieves the second-best mAP and the second-best top-1 accuracy. On the more challenging multi-view gallery of PRW, the proposed model also obtain the second-best mAP and the best top-1 accuracy. We observe that \cite{coat,pstr} construct sophisticated attention blocks \cite{vit,ddetr} and multi-scale feature representations, which can be further incorporated to boost the performance of our proposed model. When combining our proposed method with  \cite{coat}, our proposed model achieves the best mAP and the second best top-1 accuracy.

\begin{table}[h]
	\centering
	\begin{tabular}{l|ccc}
		\toprule
		\multirow{2}{*}{\textbf{Methods}} & \textbf{Training} & \textbf{Params} & \textbf{Run time} \\
		& \textbf{Time (h)} & \textbf{(M)} & \textbf{(ms)}\\
		\midrule
		HOIM \cite{hoim} & 5.9 & 33.5 & 73 \\
		NAE \cite{nae} & 5.2 & 33.5 & 71 \\
		ours$_{34}$ & 2.1 + 1.7 & 42.9 & 35 \\
		ours & 3.3 + 2.7 & 56.3 & 64 \\
		\midrule
		DMRNet \cite{dmrnet} & - & 49.1 & - \\
		ours$_{34}$ (Retina) & 3.0 + 1.7 & 50.9 & 42 \\
		ours (RetinaNet) & 4.6 + 2.7 & 59.6 & 58 \\
		\midrule
		AlignPS \cite{alignps} & 19.0 & 42.2 & 51\\
		AlignPS+ \cite{alignps} & 20.7 & 43.1 & 54\\
		ours$_{34}$(FCOS) & 2.9 + 1.7 & 50.2 & 41\\
		ours(FCOS) & 4.4 + 2.7 & 55.3 & 56\\
		\bottomrule
	\end{tabular}
	
	\caption{Efficiency comparison between the proposed fully decoupled models and previous decoupled models. The result of \cite{dmrnet} is estimated according to the paper as the source code is not released yet. }
	\label{tab:time}
	
\end{table}

\textbf{Efficiency comparison}. To verify the efficiency of the fully decoupled person search, we compare the proposed models with previous decoupled models by training time (h), number of parameters (M), and run time (ms), in Table \ref{tab:time}. All results are conducted on the PRW dataset by a single RTX 3090 GPU. It can be observed that the proposed models are with more parameters. Yet the inference time is barely increased due to the highly parallel model architecture. Although the task-incremental person search network introduces another training stage, the fully decoupled mechanism makes the training of both sub-tasks converge faster, resulting in an acceptable total training time.

\section{Conclusion and Limitations}\label{conc}
Decoupling end-to-end person search has been effectively explored for end-to-end person search to date. By analyzing their respective limitations, this paper takes a further step to enable fully decoupled end-to-end person search which achieves the optimum for both sub-tasks towards optimal person search performance. Notably, we design a task-incremental person search framework that decouples the end-to-end model architecture for the contradictory sub-tasks. The proposed task-incremental person search network further enables task-incremental training for the two sub-tasks, leading to fully decoupled end-to-end person search. Experimental results demonstrate the advantages of our proposed fully decoupled person search models. Moreover, this paper presents only baseline models for fully decoupled person search. We shall explore improving the parameter-efficiency and effectiveness of the proposed fully decoupled person search mechanism in future research.

\end{document}